\documentclass{article}

\usepackage{microtype}
\usepackage{graphicx}
\usepackage{subcaption}
\usepackage{multirow}
\usepackage{pifont}
\newcommand{\cmark}{\ding{51}}
\newcommand{\xmark}{\ding{55}}
\usepackage{booktabs} 
\usepackage{dsfont} 

\usepackage{hyperref}

\usepackage[accepted]{icml2026}

\usepackage{amsmath}
\usepackage{amssymb}
\usepackage{mathtools}
\usepackage{amsthm}

\usepackage[capitalize,noabbrev]{cleveref}

\theoremstyle{plain}

\theoremstyle{definition}

\theoremstyle{remark}

\usepackage{comment}

\usepackage[textsize=tiny]{todonotes}

\icmltitlerunning{Auditing and Filtering Modality Collapse in Traffic VideoQA}

\begin{document}

\twocolumn[
  \icmltitle{From Accuracy to Visual Dependence: Auditing and Filtering Modality Collapse in Traffic VideoQA}



  \icmlsetsymbol{equal}{*}
\begin{icmlauthorlist}
    \icmlauthor{Sena Korkut}{tum}
    \icmlauthor{María Alejandra Bravo Sarmiento}{tum,helmholtz,mcml}
    \icmlauthor{Sanghwan Kim}{tum,helmholtz,mcml}
    \icmlauthor{Zeynep Akata}{tum,helmholtz,mcml}
  \end{icmlauthorlist}

  \icmlaffiliation{helmholtz}{Helmholtz Zentrum München, Munich, Germany}
  \icmlaffiliation{tum}{Technical University of Munich, Munich, Germany}
  \icmlaffiliation{mcml}{Munich Center for Machine Learning, Munich, Germany}

  \icmlcorrespondingauthor{Sena Korkut}{sena.korkut@tum.de}
  \icmlkeywords{Machine Learning, ICML}

  \vskip 0.3in
]



\printAffiliationsAndNotice{}  
\begin{abstract}
High benchmark accuracy does not guarantee genuine use of visual evidence. We study this problem in traffic accident Video Question Answering (VideoQA), where correct answers should depend on scene-specific visual evidence but may instead be inferred from textual shortcuts. Through an audit of four public benchmarks, we find that several recent open-weight Vision-Language Models (VLMs) perform competitively, and sometimes better, without video input. On the MM-AU benchmark, removing video consistently improves accuracy, and adding more frames further degrades performance. To quantify visual dependence, we introduce two dataset-level diagnostics: Blind Gap, measuring above-chance text-only performance, and Visual Gain, measuring the marginal benefit of adding video. We further propose an instance-level Shortcut Score that combines text-only confidence with visual necessity signals, enabling continuous, training-free filtering of shortcut-prone questions. The resulting subsets reduce shortcut bias and improve visual grounding. Our findings reveal large differences in grounding quality across benchmarks and show that visually grounded evaluation, not just high accuracy, is essential in safety-critical VideoQA.
\end{abstract}

\begin{figure}[t]
    \centering
    \includegraphics[width=0.9\linewidth]{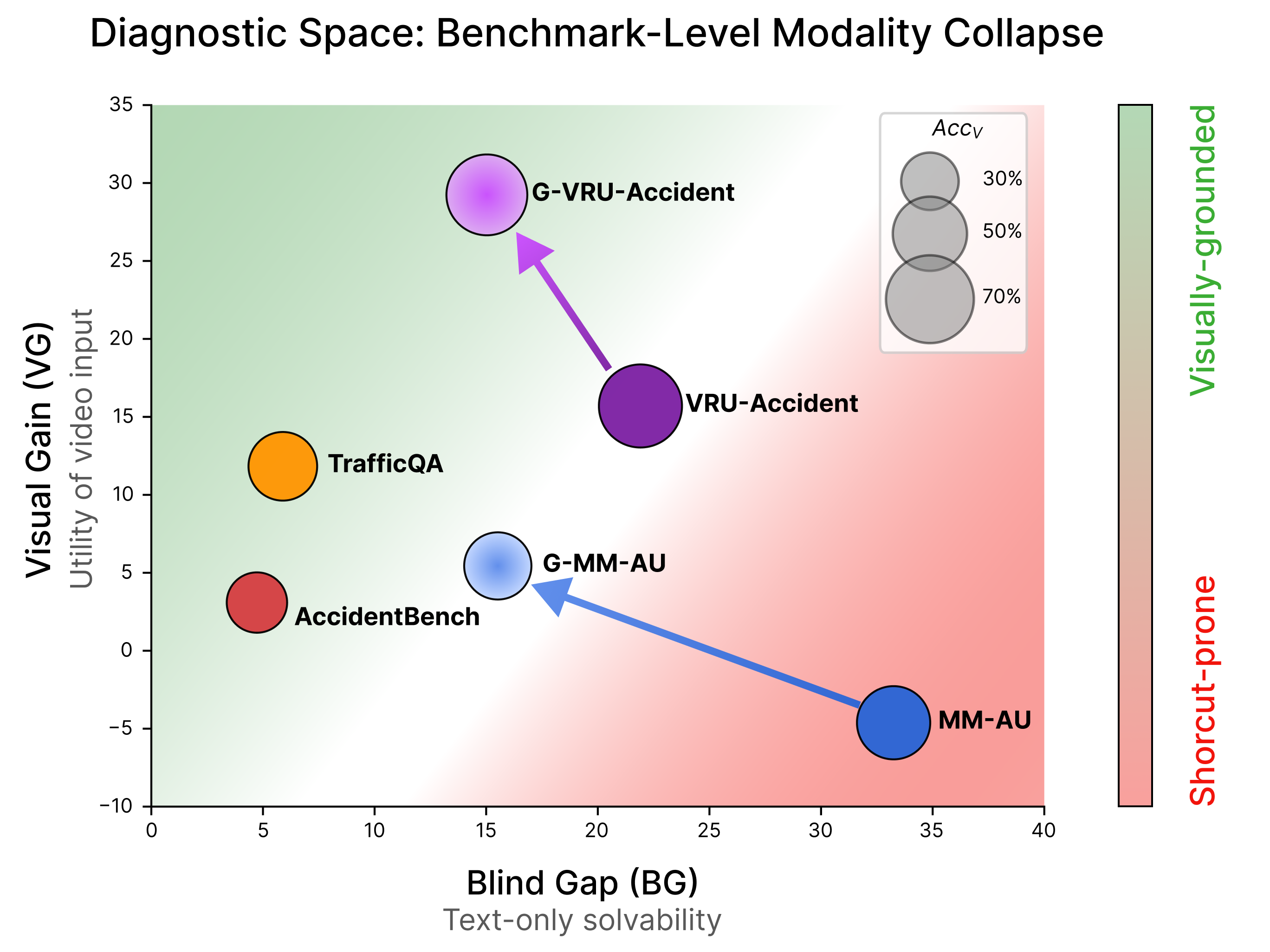}
    \caption{Diagnostic space of benchmark-level modality collapse. Each benchmark is positioned by its Blind Gap and Visual Gain, with circle size proportional to multimodal accuracy ($Acc_V$). 
    Our shortcut-aware filtering shifts MM-AU and VRU-Accident toward lower Blind Gap and higher Visual Gain, producing more visually grounded evaluation subsets (G-MM-AU and G-VRU-Accident).
}
    \label{fig:teaser}
    \vspace{-0.4cm}
\end{figure}

\section{Introduction}
Benchmarks guide research priorities and serve as the primary basis for measuring progress. Yet benchmark gains do not always reflect genuine advances in the intended capability. In multimodal learning, models may answer correctly while ignoring visual input and relying instead on language priors or dataset-specific shortcuts~\citep{goyal2017makingvvqamatter,agrawal2018dont,dancette2021beyond}. This is especially concerning in traffic accident VideoQA, where correct answers should depend on scene-specific evidence such as interacting vehicles, pedestrians, temporal dynamics, and causal structure.

We study this failure mode as \emph{modality collapse}: the visual modality contributes little or no useful information despite the task nominally requiring visual reasoning~\citep{sim2025,deng2025words}. Across several traffic accident benchmarks, recent open-weight VLMs perform competitively, and sometimes better, without video input. On MM-AU, removing video consistently improves accuracy, and adding more frames further degrades performance. In contrast, SUTD-TrafficQA and VRU-Accident show measurable gains from video input. These results suggest that benchmark accuracy alone cannot distinguish genuine visual reasoning from shortcut exploitation.

To evaluate visual dependence, we introduce a lightweight diagnostic framework based on two dataset-level metrics: \emph{Blind Gap}, which measures how far text-only accuracy exceeds chance, and \emph{Visual Gain}, which measures the benefit of adding video input. We further introduce a continuous instance-level \emph{Shortcut Score} that combines confidence-weighted text-only evidence with a visual necessity signal, enabling shortcut-aware filtering without auxiliary training.

We audit four public traffic accident VideoQA benchmarks: MM-AU, VRU-Accident, SUTD-TrafficQA, and AccidentBench. As shown in Figure~\ref{fig:teaser}, the benchmarks occupy very different regions in the Blind Gap–Visual Gain space despite comparable accuracy, showing that benchmark accuracy alone does not reflect visual grounding quality. MM-AU lies in the shortcut-prone region, with high Blind Gap and negative Visual Gain, while TrafficQA shows lower Blind Gap and consistently positive Visual Gain. Shortcut-aware filtering shifts MM-AU and VRU-Accident toward the visually grounded region by reducing text-only solvability and increasing the utility of video input. These findings highlight that, in safety-critical VideoQA, systems must succeed because they use visual evidence, not because they exploit annotation artifacts.

\section{Related Work}
VideoQA benchmarks have expanded multimodal evaluation to temporal and causal
reasoning, yet benchmark accuracy alone cannot verify genuine visual
grounding~\citep{chen2024mmstar,zhang2026vidground}. VLMs inherit strong language
priors from pretraining that persist even as visual capacity scales, pointing to a
utilization rather than capacity bottleneck~\citep{ghosh2026dovlms,long2026languageprior}.
Shortcut learning~\citep{geirhos2020shortcut} in VQA has motivated binary filtering
approaches~\citep{asadi2026mirage,zhang2026vidground} and dataset-level
auditing metrics~\citep{chen2024mmstar,zafar2026grounding}, but neither identifies
shortcut-prone questions at the instance level with a continuous score. Our work
provides the first such audit along with a filtering framework across traffic accident VideoQA benchmarks.
See Appendix~\ref{app:related_work} for a full discussion.

\section{Diagnosing and Filtering Modality Collapse}

A VideoQA benchmark should require visual evidence to answer correctly. We define a question as \emph{visually grounded} if video input improves prediction toward the correct answer, and \emph{shortcut-prone} if it can be solved from text alone. We diagnose this at both the dataset and question levels to construct grounded evaluation subsets.

\subsection{Stage 1: Dataset-Level Diagnostics}
\label{sec:stage1}

We evaluate models in two settings: \emph{blind}, where only the question and answer options are provided, and \emph{video}, where uniformly sampled frames are added. The \textbf{Blind Gap} (BG) measures how far text-only accuracy exceeds random chance:
\begin{equation}
    \text{BlindGap}(\mathcal{D}) = \text{Acc}_{\text{blind}} - \text{Acc}_{\text{random}},
    \label{eq:blind_gap}
\end{equation}
where $\text{Acc}_{\text{random}} = 1/K$ for a $K$-option question. The \textbf{Visual Gain} (VG) measures the accuracy change from adding video:
\begin{equation}
    \text{VisualGain}(\mathcal{D}) = \text{Acc}_{\text{video}} - \text{Acc}_{\text{blind}}.
    \label{eq:visual_gain}
\end{equation}
A high BG indicates text-only solvability, while positive VG indicates useful visual evidence. Together, high BG with low or negative VG signals shortcut exploitation, whereas low BG with positive VG indicates stronger visual grounding.

\subsection{Stage 2: Question-Level Shortcut Diagnostics}
\label{sec:stage2}

Dataset-level averages do not reveal which questions drive shortcut behavior. We therefore define a continuous per-question \textbf{Shortcut Score} $S(q)$ from textual evidence and visual necessity.

The \textbf{Textual Evidence Score} $T(q)$ measures whether text-only LLMs answer correctly and confidently. For each model $m \in \mathcal{M}_\text{LLM}$, let $H_m(q) = -\sum_{k=1}^{K} p_{m,k}(q)\log p_{m,k}(q)$ be the entropy of the answer distribution and $H_{\text{norm},m}(q)=H_m(q)/\ln K \in [0,1]$. Only correct blind predictions contribute:
\begin{equation}
    T(q) = \frac{1}{|\mathcal{M}_\text{LLM}|} \sum_{m \in \mathcal{M}_\text{LLM}} [\text{Correct}_{m}(q)] \cdot \bigl(1 - H_{\text{norm},m}(q)\bigr).
    \label{eq:T}
\end{equation}
High $T(q)$ indicates a reliable linguistic shortcut.

The \textbf{Visual Necessity Score} $V(q)$ measures whether adding video increases the probability assigned to the correct answer. For each VLM, we compute:
\begin{equation}
\label{eq:delta}
\Delta_m(q) =
P_m(y^\star \mid \text{video}, q)
-
P_m(y^\star \mid \text{blind}, q),
\end{equation}
and average the normalized gain across models:
\begin{equation}
\label{eq:V}
V(q) =
\frac{1}{|\mathcal{M}_{\text{VLM}}|}
\sum_{m \in \mathcal{M}_{\text{VLM}}}
\frac{\Delta_m(q)}{1 - 1/K}.
\end{equation}
Here, $y^\star$ is the correct answer and $K$ is the number of answer options. Positive $V(q)$ indicates that video shifts probability toward the correct answer; near-zero or negative $V(q)$ indicates that visual input is unnecessary or harmful.

\begin{table*}[t]
    \centering
    \caption{Global diagnosis of modality collapse. We report Blind Accuracy
    ($Acc_B$), Multimodal Accuracy ($Acc_V$), Blind Gap (BG$\downarrow$), and Visual
    Gain (VG$\uparrow$) in \%. Text-only LLMs have no video accuracy by design.}
    \label{tab:global_modality_diagnosis}
    \resizebox{\textwidth}{!}{%
    \begin{tabular}{lcccccccccccccccc}
        \toprule
        & \multicolumn{4}{c}{\textbf{MM-AU}}
        & \multicolumn{4}{c}{\textbf{TrafficQA}}
        & \multicolumn{4}{c}{\textbf{Acc.Bench}}
        & \multicolumn{4}{c}{\textbf{VRU-Acc.}} \\
        \cmidrule(lr){2-5}\cmidrule(lr){6-9}\cmidrule(lr){10-13}\cmidrule(lr){14-17}
        \textbf{Model}
        & $Acc_B$  & $Acc_V$  & $BG\downarrow$ & $VG\uparrow$
        & $Acc_B$  & $Acc_V$ & $BG\downarrow$ & $VG\uparrow$
        & $Acc_B$  & $Acc_V$ & $BG\downarrow$ & $VG\uparrow$
        & $Acc_B$  & $Acc_V$ & $BG\downarrow$ & $VG\uparrow$ \\
        \midrule
        \textbf{Average VLM} 
        & 53.25 & 48.64 & 33.25 & -4.62
        & 30.89 & 42.71 & 5.89 & 11.83
        & 29.97 & 33.05 & 4.73 & 3.09
        & 46.91 & 62.61 & 21.91 & 15.70 \\
        \midrule
        InternVL 2.5 8B  & 52.14 & 50.35 & 32.14 & -1.79  & 29.72 & 41.95 & 4.72 & 12.23 & 29.64 & 31.68 & 4.40  & 2.05  & 50.18 & 63.62 & 25.18 & 13.43 \\
        InternVL 3.5 8B  & 60.33 & 50.88 & 40.33 & -9.46  & 33.62 & 45.67 & 8.62 & 12.06 & 30.43 & 37.76 & 5.19  & 7.33  & 42.95 & 62.85 & 17.95 & 19.90 \\
        Qwen 2.5 VL 7B   & 54.83 & 47.86 & 34.83 & -6.97  & 31.52 & 43.53 & 6.52 & 12.01 & 30.50 & 35.18 & 5.26  & 4.69  & 44.40 & 60.92 & 19.40 & 16.52 \\
        LLaVA-OV 8B      & 45.70 & 45.45 & 25.70 & -0.24  & 28.69 & 39.70 & 3.69 & 11.01 & 29.31 & 27.59 & 4.07  & -1.72 & 50.10 & 63.03 & 25.10 & 12.93 \\
        \midrule
        Llama 3.1 8B     & 46.60 & -     & 26.60 & -      & 30.45 & -     & 5.45 & -     & 20.33 & -     & -4.91 & -     & 42.77 & -     & 17.77 & -     \\
        Mistral 7B v0.1  & 33.92 & -     & 13.92 & -      & 28.52 & -     & 3.52 & -     & 26.60 & -     & 1.36  & -     & 31.18 & -     & 6.18  & -     \\
        Qwen 2.5 14B     & 51.45 & -     & 31.45 & -      & 28.18 & -     & 3.18 & -     & 27.00 & -     & 1.76  & -     & 44.55 & -     & 19.55 & -     \\
        \bottomrule
    \end{tabular}}
\end{table*}

\subsection{Stage 3: Shortcut-Aware Filtering}
\label{sec:stage3}

The \textbf{Shortcut Score} combines the two signals:
\begin{equation}
    S(q) = T(q) - V(q).
    \label{eq:S}
\end{equation}
Textual solvability raises $S(q)$, while visual necessity lowers it; thus high $S(q)$ indicates shortcut-prone questions, and low or negative $S(q)$ indicates visually useful questions.

We construct a grounded subset by retaining questions below a threshold $\tau$:
\begin{equation}
    \mathcal{D}_{\text{grounded}} = \{ q \in \mathcal{D} \mid S(q) \leq \tau \}.
    \label{eq:filter}
\end{equation}
The threshold controls the grounding--coverage trade-off: lower $\tau$ yields cleaner but smaller subsets, while higher $\tau$ preserves more data with more residual shortcut risk. Since $S(q)$ is computed from instance-level probabilities, filtering does not directly optimize dataset-level BG or VG.

\section{Experiments}
\textbf{Benchmarks Under Study}. 
We evaluate on four public traffic accident VideoQA benchmarks: MM-AU~\citep{fang2024mmau}, VRU-Accident~\citep{kim2025vru}, SUTD-TrafficQA~\citep{xu2021sutdtrafficqa}, and AccidentBench~\citep{gu2025accidentbench}. All use a multiple-choice format over real-world accident videos, enabling controlled blind/video evaluation. Benchmark-specific details are provided in Appendix~\ref{app:benchmark}.

\textbf{Evaluation Protocol}. 
We evaluate each VLM in two settings: \emph{blind}, where the model receives only the question and answer options, and \emph{video}, where it additionally receives uniformly sampled video frames. Unless otherwise stated, we use 16 frames per video. All tasks are evaluated as multiple-choice QA using accuracy. We report blind accuracy, video accuracy, BG, and VG; instance-level Shortcut Scores are computed from option probabilities under blind and video conditions. Further details are provided in Appendix~\ref{app:evaluation_protocol}.

\begin{figure*}[t]
    \centering
    \includegraphics[width=\textwidth]{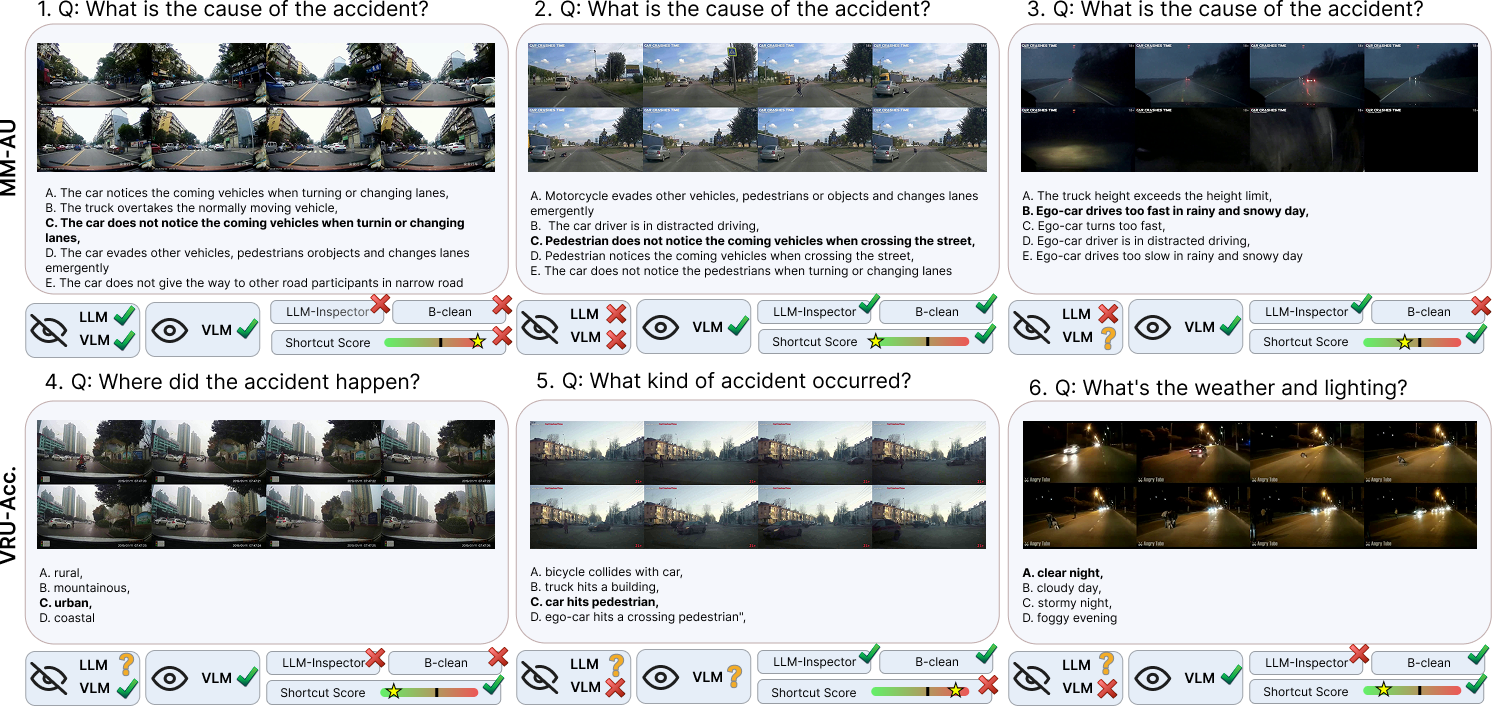}
    \caption{Representative questions from MM-AU and VRU-Accident. Masked-eye and open-eye icons show blind and video ensemble predictions (\cmark\ all-correct, \xmark\ all-incorrect, $?$ mixed). For filtering methods, \cmark/\xmark\ indicate retained/removed. The Shortcut Score bar reflects shortcut severity.}
    \vspace{-0.4cm}
    \label{fig:qualitative}
\end{figure*}

\subsection{Benchmark Audit Results}

Table~\ref{tab:global_modality_diagnosis} reports BG and VG across
four traffic accident VideoQA benchmarks and seven models. The severity varies substantially by benchmark. MM-AU shows the most extreme case: BG reaches 40.33, and VG is negative for all four VLMs, meaning that video input not only fails to help but often degrades accuracy. To test whether this is caused by sparse frame sampling, we increased the frame count from 8 to 64 and 256 for InternVL2.5-8B. Accuracy declined at each step, suggesting that the failure is structural rather than a simple sampling artifact: the benchmark can often be answered from text alone, and additional visual context does not resolve the issue. The fact that text-only LLMs match or exceed several VLMs on MM-AU further supports this interpretation, indicating that exploitable signal lies largely in the question and answer text.

VRU-Accident and TrafficQA represent different points on the same spectrum. On VRU-Accident, VG is positive and large, yet BG remains high, reaching 25 points for two models. This means that the benchmark contains useful visual signal, but also remains partially solvable from text alone. In contrast, TrafficQA has BG below 9 across all architectures and VG around 11--12 points, providing the strongest evidence of visual dependence among the traffic benchmarks studied here. AccidentBench falls between these cases, with low BG and modest VG.

This pattern is not unique to traffic accident VideoQA: Appendix~\ref{app:general_videoqa} shows that widely used general VideoQA benchmarks also exhibit non-trivial BG, even when VG remains positive.


\begin{table}[b]
\vspace{-1pt}
\centering
\small
\caption{Comparison of filtering strategies on MM-AU and VRU-Accident using
LLaVA-OneVision. We report 
video accuracy ($Acc_V$), Blind Gap (BG), Visual Gain (VG), and the number of remaining samples.}
\label{tab:filtering_comparison_llava}
\resizebox{\linewidth}{!}{%
\begin{tabular}{llcccc}
\toprule
\textbf{Dataset} & \textbf{Metric}
& \textbf{No Filtering}
& \textbf{LLM-Inspector}
& \textbf{B-Clean}
& \textbf{S-score} \\
\midrule

\multirow{4}{*}{\textbf{MM-AU}}
& $Acc_V$               & 45.45 & 31.79 & 23.54 & 37.80 \\
& BG$\downarrow$      & 25.70 & 11.47 & -5.02 & 10.38 \\
& VG$\uparrow$        & -0.24 &  0.32 & -19.91 &  7.42 \\
& \# QA          & 2453  & 1252  & 701   & 836 \\
\midrule
\multirow{4}{*}{\shortstack{\textbf{VRU-}\\\textbf{Accident}}}
& $Acc_V$               & 63.03 & 52.80 & 52.38 & 69.25 \\
& BG$\downarrow$      & 25.10 & 12.16 & -3.22 & 18.76 \\
& VG$\uparrow$        & 12.93 & 15.64 & -10.65 & 25.49 \\
& \# QA          & 6000  & 3926  & 2709  & 4088 \\
\bottomrule

\end{tabular}}
\end{table}

\begin{figure}[t]
    \centering
    \includegraphics[width=0.9\linewidth]{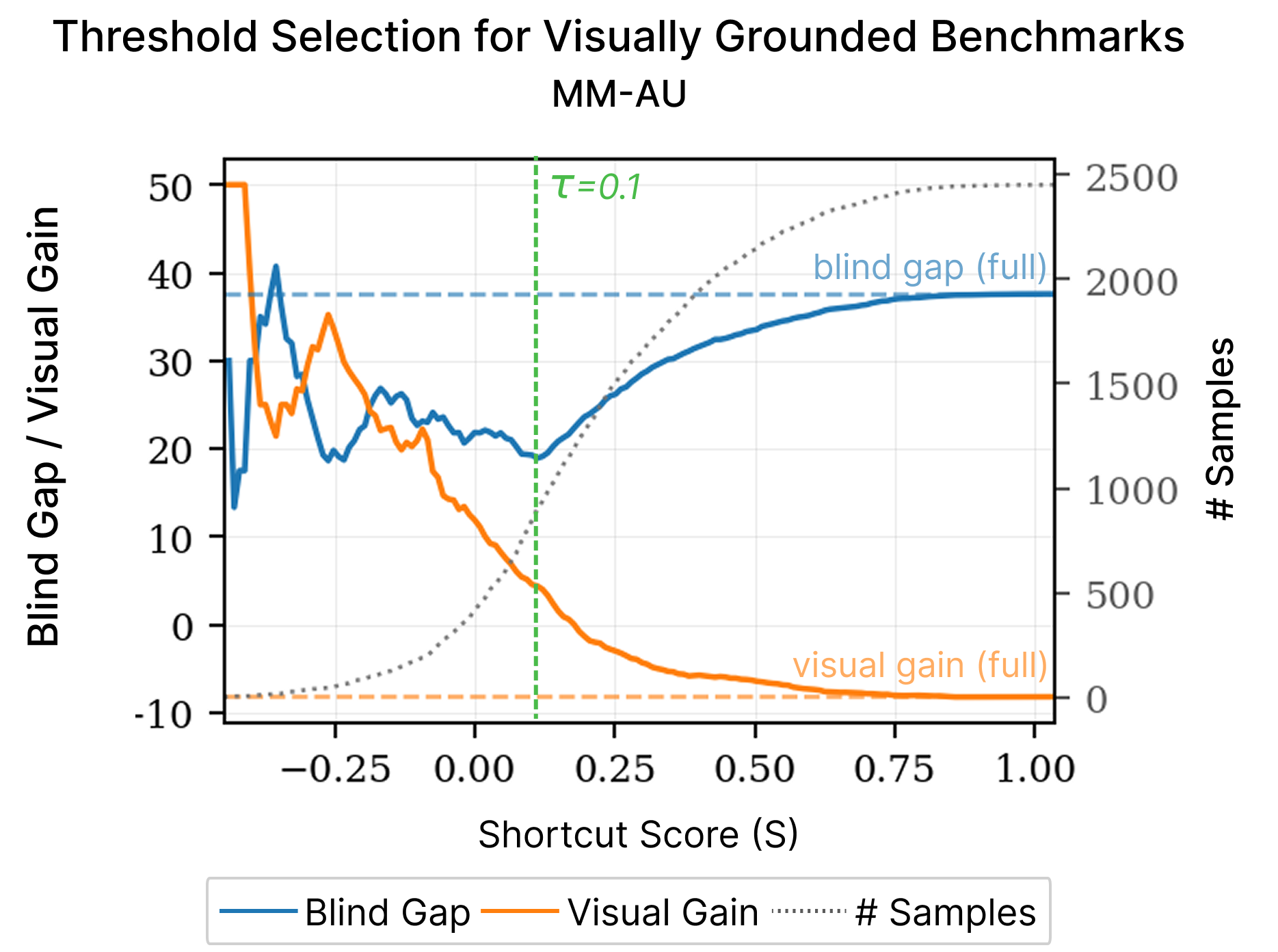}
    \caption{
    Sweeping the Shortcut Score threshold $\tau$ reveals the tradeoff between BG, VG, and the number of retained samples. Choosing $\tau = 0.1$ reduces text-only solvability while improving visual grounding.}
    \label{fig:threshold_selection}
    \vspace{-0.4cm}
\end{figure}

\subsection{Filtering Results}
Figure~\ref{fig:threshold_selection} shows the $S$ sweep on MM-AU: as $\tau$
decreases, BG drops below its full-dataset level and VG turns positive, with $\tau = 0.1$ marking a stable point before sample count becomes too small for reliable estimates. We additionally sweep $\tau$ over $T$ and $-V$ to understand each signal's contribution: filtering on $T$ reduces BG but cannot recover positive VG on MM-AU; filtering on $-V$ improves VG but raises BG and collapses sample size. The $S$ sweep provides the most balanced outcome on both datasets, and we apply $\tau = 0.1$ to both MM-AU and VRU-Accident. Full sweep results across all three signals and both datasets are provided in Appendix~\ref{app:filtering}.

We compare S-score against two binary filtering baselines using LLaVA-OneVision-8B, held out from all filtering decisions (Table~\ref{tab:filtering_comparison_llava}). LLM-Inspector~\citep{chen2024mmstar} removes questions answered correctly by a text-only LLM panel; B-Clean~\citep{asadi2026mirage} removes questions answered correctly in blind mode. Both reduce BG, but B-Clean yields strongly negative VG on both datasets ($-19.91$ on MM-AU, $-10.65$ on VRU-Accident), indicating that its binary rule removes visually useful questions alongside shortcut-prone ones. LLM-Inspector retains positive VG but has no mechanism to verify that video
actually helps on retained questions. S-score achieves the best trade-off: on MM-AU, BG drops from $25.70$ to $10.38$ and VG shifts from $-0.24$ to $7.42$, a structural reversal confirmed by text-only LLMs falling to near or below random chance on the filtered subset (see Appendix~\ref{app:filtering}). On VRU-Accident, BG is roughly halved and VG nearly doubles to $25.49$. This is because S-score retains questions where video shifts model behavior, even when the blind answer was correct close to random chance, which is a distinction that binary filters cannot make. Figure~\ref{fig:qualitative} illustrates this with representative examples, showing cases where LLM-Inspector and B-Clean disagree with S-score and where the Shortcut Score bar aligns with the filtering decision.

Looking at the first two MM-AU examples in Figure~\ref{fig:qualitative} reveals how repetitive text patterns drive these shortcuts. In the first example, all models predict the correct answer (C) even without the video, showing a strong text bias. In the second example, all blind models incorrectly choose option E, which has a very similar sentence structure to the shortcut in the first instance. This phrase is highly repetitive; in fact, this exact answer structure is the ground truth 234 times across the MM-AU dataset. This shows that the modality collapse at dataset level is also closely linked to a lack of answer diversity, which allows models to simply exploit high-frequency shortcut patterns rather than look at the video.

We additionally evaluate fine-tuned variants of Qwen2.5-VL-7B on both standard and grounded subsets to assess how training interacts with BG and VG structure; results and discussion are provided in Appendix~\ref{app:training}.

\section{Conclusion}
We introduce Blind Gap, Visual Gain, and a Shortcut Score as lightweight diagnostics for quantifying modality collapse and identifying shortcut-prone questions in VideoQA benchmarks. Our results show that several traffic accident benchmarks can often be solved without visual input, while shortcut-aware filtering produces more visually grounded evaluation subsets. These findings highlight the need for future multimodal benchmarks to measure not only accuracy, but also genuine visual dependence. Furthermore, our analysis offers an insight for future dataset construction: avoiding repetitive question formats and identical answer phrasing is vital, as common text patterns allow models to easily exploit shortcuts.

\bibliography{example_paper}
\bibliographystyle{icml2026}

\newpage
\appendix
\onecolumn
\section{Related Work}\label{app:related_work}

\textbf{VideoQA Benchmarks and Traffic Accident Understanding.} 
VideoQA benchmarks~\citep{xiao2021nextqanextphasequestionansweringexplaining,wu2024starbenchmarksituatedreasoning,mangalam2023egoschemadiagnosticbenchmarklongform,fu2025videomme,wu2024longvideobench,zhou2025mlvu} have expanded multimodal evaluation to richer temporal and causal reasoning over video, with the implicit assumption that such tasks require genuine visual grounding, which accuracy alone cannot verify~\citep{chen2024mmstar,zhang2026vidground}. Traffic accident VideoQA is a critical test case: accident causality, agent behavior, and collision dynamics are inherently scene-specific and should not be answerable from question text or commonsense alone~\citep{zhang2025seeunsafe}, yet whether existing benchmarks enforce this requirement has never been tested. SUTD-TrafficQA, MM-AU, VRU-Accident, and AccidentBench~\citep{xu2021sutdtrafficqa,fang2024mmau,kim2025vru,gu2025accidentbench} target precisely this reasoning but report only answer accuracy and no blind baselines, which is the gap this work addresses.

\textbf{Vision-Language Models and the Roots of Language Dominance.} 
Modern VLMs combine a visual encoder, projection module, and autoregressive language backbone, extended to video via frame sampling and spatiotemporal encodings~\citep{bai2025qwen25vltechnicalreport,chen2024internvl,li2024llavaov}. Because language backbones are pretrained on far larger text corpora than the image-text or video-text data used for visual alignment~\citep{alayrac2022flamingo,liu2023llava}, models inherit strong language priors before acquiring robust visual grounding. Increasing visual capacity through higher-resolution inputs or denser sampling does not resolve this: image-token representations stabilize early and remain accurate under aggressive visual truncation~\citep{ghosh2026dovlms}, pointing to a utilization rather than capacity bottleneck. Mechanistic studies confirm this view: visual influence emerges only after a late integration point~\citep{long2026languageprior}, and language priors can dominate predictions even when visual attributes are encoded~\citep{nooralahzadeh2026arbitration}.

\textbf{Shortcut Learning, Modality Collapse, and Benchmark Auditing.}
Shortcut learning~\citep{geirhos2020shortcut} in VQA motivated bias-controlled
datasets~\citep{goyal2017makingvvqamatter,agrawal2018dont} and was later extended to joint
image-question correlations~\citep{dancette2021beyond}. In multimodal systems
this manifests as \textit{modality collapse}, where language dominates prediction
regardless of visual input~\citep{sim2025,deng2025words}. In traffic accident
VideoQA, DriveBench~\citep{xie2025} confirms this by showing comparable
performance under text-only and full-input conditions. Existing auditing
approaches either apply binary filtering by removing questions a blind model answers correctly~\citep{asadi2026mirage,zhang2026vidground} , or introduce
dataset-level metrics to quantify visual
contribution~\citep{chen2024mmstar,zafar2026grounding,brown2025tst,lee2025retina}.
Binary filtering discards visually ambiguous but grounded questions; dataset-level
metrics cannot identify which individual questions are shortcut-prone. Our
\emph{Shortcut Score} addresses both limitations by assigning each question a
continuous value that combines confidence-weighted text-only evidence with a
visual necessity signal, and we provide the first such audit across traffic
accident VideoQA benchmarks.

\section{Experimental Extension}

\subsection{Benchmarks Under Study}  \label{app:benchmark}

\begin{table}[t]
    \centering
    \small
    \caption{Comparison of the traffic accident VideoQA benchmarks analyzed in this work.}
    \label{tab:dataset_summary}
    \resizebox{\columnwidth}{!}{%
    \begin{tabular}{lcccccl}
        \toprule
        \textbf{Dataset} & \textbf{Videos} & \textbf{Avg.\ \# Frames} 
        & \textbf{QA Pairs} & \textbf{Choices} & \textbf{Format} 
        & \textbf{Reasoning Types} \\
        \midrule
        MM-AU        & 11,727 & 187  & 11,727 & 5      & MCQ & Causal \\
        VRU-Accident & 1,000  & 189  & 6,000  & 4      & MCQ 
            & Causal, counterfactual, scene attr. \\
        TrafficQA    & 10,080 & 189  & 62,535 & 4      & MCQ 
            & 6 types incl.\ counterfactual \\
        AccidentBench & 2,000 & 1330 & 17,069 & 2--12  & MCQ 
            & Temporal, spatial, intent \\
        \bottomrule
    \end{tabular}}
\end{table}

We select four publicly available traffic accident VideoQA benchmarks, and the summary is provided in
Table~\ref{tab:dataset_summary}.

\textbf{MM-AU} \citep{fang2024mmau} covers 58 accident categories in ego-view
dashcam footage. Every question asks the same causal query:``What is the cause
of the accident?'', with distractors drawn from other accident categories. This homogeneous question distribution makes it particularly susceptible to
language-prior exploitation.

\textbf{VRU-Accident} \citep{kim2025vru} focuses on accidents involving pedestrians and cyclists across six question categories. Distractors are generated
by GPT-4o conditioned on the correct answer and video context, producing semantically plausible options. The large answer pool, 712 unique prevention
methods and 437 unique accident reasons, limits frequency-based shortcuts.

\textbf{SUTD-TrafficQA} \citep{xu2021sutdtrafficqa} provides human-annotated QA pairs
across six reasoning types. Answer candidates are sampled to reduce repetition,
and the authors report text-only baselines as a diagnostic reference. The
counterfactual and reverse reasoning splits are especially relevant for visual
grounding, as they concern outcomes unlikely to be resolved from language priors alone.

\textbf{AccidentBench} \citep{gu2025accidentbench} targets temporal, spatial, and
intent reasoning with a tiered answer set that modulates difficulty. Videos can
reach approximately 16 minutes, posing a practical challenge for models under a
fixed frame budget. We use only the Land domain subset.

\begin{figure*}[t]
    \centering
    \includegraphics[width=\textwidth]{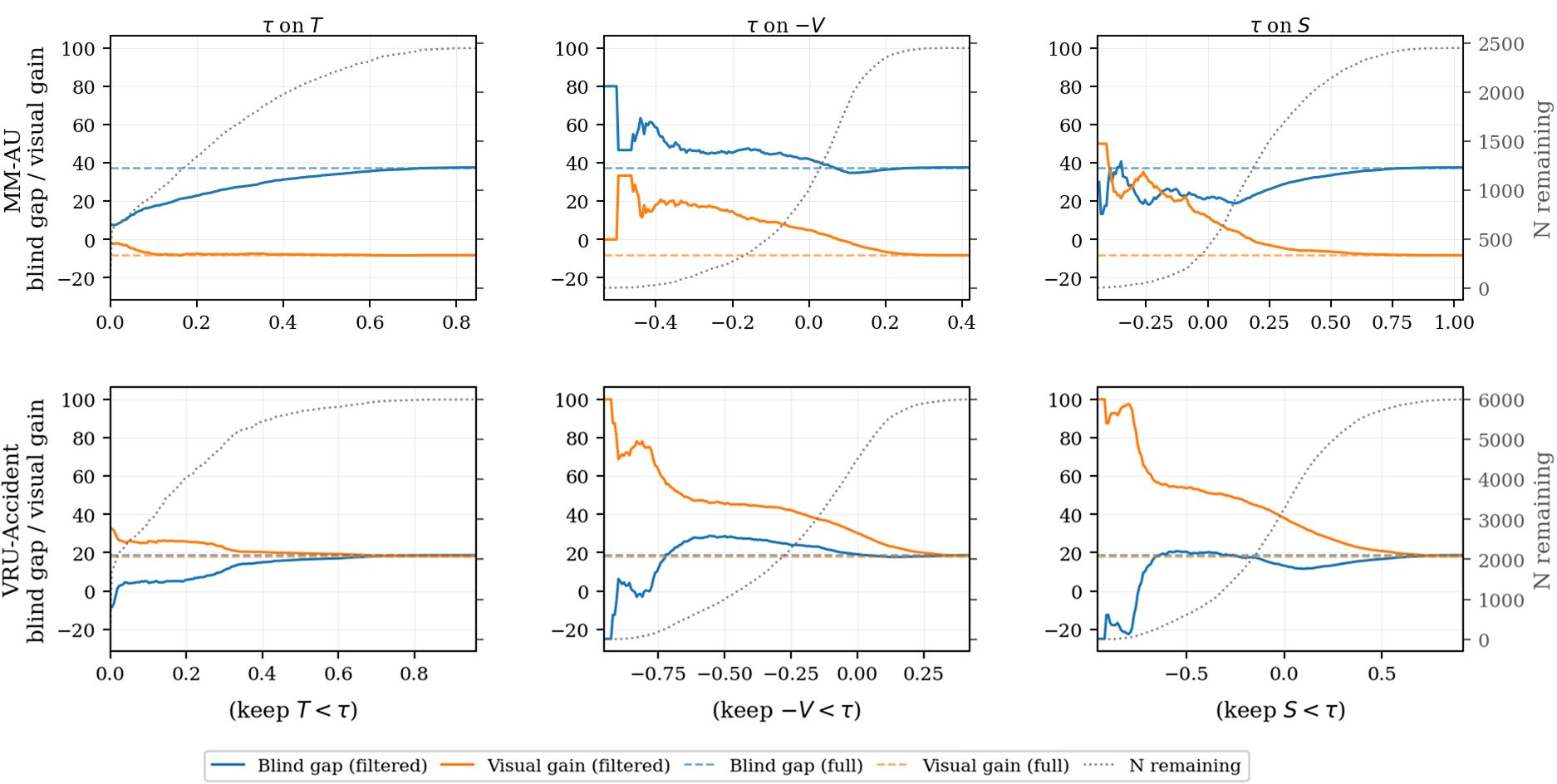}
    \caption{Threshold sweep on $T$, $-V$, and $S$ for MM-AU and VRU-Accident. Blue and orange curves show Blind Gap and Visual Gain on the filtered subset; dashed lines mark the full-dataset values. The gray dotted curve shows the number of remaining samples on the right axis.}
    \label{fig:sweeps}
\end{figure*}

\subsection{Evaluation Protocol} \label{app:evaluation_protocol}

We evaluate four vision-language models at the 7--8B parameter scale:
Qwen2.5-VL-7B, InternVL2.5-8B, InternVL3.5-8B, and LLaVA-OneVision.
To isolate the contribution of linguistic patterns, we include three text-only LLM baselines that receive no visual input: Llama 3.1-8B, Mistral-7B v0.1, and Qwen 2.5-14B.

All models are evaluated in two settings: blind (question and answer options only, no video) and video (full multimodal input with 16 uniformly sampled frames). Accuracy is the primary metric. Random chance baselines are dataset-specific: 25\% for 4-option questions (VRU-Accident, TrafficQA), 20\% for 5-option questions (MM-AU), and 25.24\% on average for AccidentBench.

\textbf{Shortcut Score.} Instance-level scores are computed using two VLMs, Qwen2.5-VL-7B and InternVL3.5-8B, evaluated in both blind and video mode to compute $V(q)$, and three text-only LLMs, Qwen 2.5-14B, Llama 3.1-8B, and Mistral 7B v0.1, evaluated in blind mode to compute $T(q)$. These models are a subset of the evaluation panel. Answer option
probabilities are obtained via log-likelihood scoring: each answer option is scored independently by computing the log-likelihood of the option text given the question context, and the resulting scores are normalized across options to produce a probability distribution.

\textbf{Baseline filtering methods.} LLaVA-OneVision-8B is held out from all filtering pipelines and used only for evaluation. LLM-Inspector is adapted using four text-only LLMs: Qwen 2.5 14B, Llama 3.1 8B, Mistral 7B v0.1, and DeepSeek-LLM-7B-Chat. A question is discarded if more than one model answers it correctly, following the majority-based filtering criterion of the original method. B-Clean uses InternVL3.5-8B and Qwen2.5-VL-7B evaluated in blind mode, and a question is discarded if either model answers it correctly under this condition.

All experiments are run on NVIDIA GH200 GPUs. Models are prompted with a
standardized template instructing them to respond with a single option letter. Output probabilities over option tokens are extracted to compute normalized entropy and probability gain for the instance-level scores.

\begin{table*}[t]
\centering
\small
\caption{Blind and video accuracy on general VideoQA benchmarks. BG is computed
relative to random chance: 21.12\% for LongVideoBench, 25\% for VideoMME and MLVU.}
\label{tab:general_benchmarks}
\resizebox{\textwidth}{!}{%
\begin{tabular}{lcccccccccccc}
\toprule
& \multicolumn{4}{c}{\textbf{LongVideoBench}}
& \multicolumn{4}{c}{\textbf{VideoMME}}
& \multicolumn{4}{c}{\textbf{MLVU}} \\
\cmidrule(lr){2-5} \cmidrule(lr){6-9} \cmidrule(lr){10-13}
\textbf{Model}
& $Acc_B$ & $Acc_V$ & $BG\downarrow$ & $VG\uparrow$
& $Acc_B$ & $Acc_V$ & $BG\downarrow$ & $VG\uparrow$
& $Acc_B$ & $Acc_V$ & $BG\downarrow$ & $VG\uparrow$ \\
\midrule
InternVL 2.5 8B & 42.86 & 57.97 & 21.74 & 15.11 & 42.70 & 64.00 & 17.70 & 21.30 & 42.28 & 65.69 & 17.28 & 23.41 \\
InternVL 3.5 8B & 43.53 & 59.84 & 22.41 & 16.31 & 43.77 & 64.52 & 18.77 & 20.75 & 41.77 & 68.06 & 16.77 & 26.29 \\
Qwen 2.5 VL 7B  & 39.34 & 53.03 & 18.22 & 13.69 & 39.07 & 53.59 & 14.07 & 14.52 & 42.96 & 54.52 & 17.96 & 11.56 \\
\bottomrule
\end{tabular}}
\end{table*}

\begin{table*}[t]
\centering
\small
\caption{Blind accuracy, video accuracy, Blind Gap, and Visual Gain before and
after filtering for MM-AU and VRU-Accident.}
\label{tab:dataset_results_filtered}
\resizebox{\textwidth}{!}{%
\begin{tabular}{lcccccccccccccccc}
\toprule
& \multicolumn{8}{c}{\textbf{MM-AU}}
& \multicolumn{8}{c}{\textbf{VRU-Accident}} \\
\cmidrule(lr){2-9}\cmidrule(lr){10-17}
& \multicolumn{4}{c}{\textbf{Original (2453 samples)}}
& \multicolumn{4}{c}{\textbf{Filtered (G) (836 samples)}}
& \multicolumn{4}{c}{\textbf{Original (6000 samples)}}
& \multicolumn{4}{c}{\textbf{Filtered (G) (4088 samples)}} \\
\cmidrule(lr){2-5}\cmidrule(lr){6-9}\cmidrule(lr){10-13}\cmidrule(lr){14-17}
\textbf{Model}
& $Acc_B$ & $Acc_V$ & $BG\downarrow$ & $VG\uparrow$
& $Acc_B$ & $Acc_V$ & $BG\downarrow$ & $VG\uparrow$
& $Acc_B$ & $Acc_V$ & $BG\downarrow$ & $VG\uparrow$
& $Acc_B$ & $Acc_V$ & $BG\downarrow$ & $VG\uparrow$ \\
\midrule
InternVL2.5  & 52.14 & 50.35 & 32.14 & -1.79 & 33.13 & 38.28 & 13.13 & 5.14  & 50.18 & 63.62 & 25.18 & 13.43 & 43.25 & 68.81 & 18.25 & 25.56 \\
InternVL3.5  & 60.33 & 50.88 & 40.33 & -9.46 & 39.23 & 41.99 & 19.23 & 2.75  & 42.95 & 62.85 & 17.95 & 19.90 & 35.91 & 71.65 & 10.91 & 35.74 \\
LLaVA-OV     & 45.70 & 45.45 & 25.70 & -0.24 & 30.38 & 37.80 & 10.38 & 7.42  & 50.10 & 63.03 & 25.10 & 12.93 & 43.76 & 69.25 & 18.76 & 25.49 \\
Qwen2.5-VL   & 54.83 & 47.86 & 34.83 & -6.97 & 39.35 & 45.81 & 19.35 & 6.46  & 44.40 & 60.92 & 19.40 & 16.52 & 37.21 & 67.39 & 12.21 & 30.19 \\
\midrule
Llama3.1-8B  & 46.60 & --    & 26.60 & --    & 19.50 & --    & -0.50 & --    & 42.77 & --    & 17.77 & --    & 37.92 & --    & 12.92 & --    \\
Mistral-7B   & 33.92 & --    & 13.92 & --    & 11.60 & --    & -8.40 & --    & 31.18 & --    & 6.18  & --    & 20.40 & --    & -4.60 & --    \\
Qwen2.5-14B  & 51.45 & --    & 31.45 & --    & 18.66 & --    & -1.34 & --    & 44.55 & --    & 19.55 & --    & 30.94 & --    & 5.94  & --    \\
\bottomrule
\end{tabular}}
\end{table*}

\subsection{General VideoQA Benchmarks} \label{app:general_videoqa}

Table~\ref{tab:general_benchmarks} shows that this is not a problem specific to traffic datasets. On LongVideoBench \citep{wu2024longvideobench}, VideoMME, \citep{fu2025videomme} and MLVU \citep{zhou2025mlvu}, which are widely used benchmarks for video understanding, Blind Gap ranges from
roughly 14 to 22 points, comparable to VRU-Accident. A substantial share
of what these benchmarks measure is answerable without any video. Visual Gain is
positive across all models, which confirms that the visual signal contributes, but
that contribution is measured on top of an inflated baseline. The consequence is
that raw multimodal accuracy conflates two distinct effects: what the model learns
from the video, and what it would have predicted from text alone. Without
decomposing these, benchmark scores cannot be interpreted as a reliable measure of
visual reasoning ability.

\begin{table}[t]
    \centering
    \small
    \caption{Performance of Qwen2.5-VL-7B across different model configurations in (\%), reported as
    video accuracy ($Acc_V$), Blind Gap (BG), and Visual Gain (VG). Standard and Grounded variants 
    of MM-AU and VRU-Accident are shown side by side.}
    \resizebox{\textwidth}{!}{%
    \begin{tabular}{l  ccc ccc  ccc ccc  ccc  ccc}
        \toprule
        \multirow{3}{*}{\textbf{Model}}
            & \multicolumn{6}{c}{\textbf{MM-AU}}
            & \multicolumn{6}{c}{\textbf{VRU-Accident}}
            & \multicolumn{3}{c}{\textbf{TrafficQA}}
            & \multicolumn{3}{c}{\textbf{AccidentBench}} \\
        \cmidrule(lr){2-7} \cmidrule(lr){8-13} \cmidrule(lr){14-16} \cmidrule(lr){17-19}
            & \multicolumn{3}{c}{Standard}
            & \multicolumn{3}{c}{Grounded}
            & \multicolumn{3}{c}{Standard}
            & \multicolumn{3}{c}{Grounded}
            & & & & & & \\
        \cmidrule(lr){2-4} \cmidrule(lr){5-7} \cmidrule(lr){8-10} \cmidrule(lr){11-13}
            & $Acc_V$ & \textbf{BG}$\downarrow$ & \textbf{VG}$\uparrow$
            & $Acc_V$ & \textbf{BG}$\downarrow$ & \textbf{VG}$\uparrow$
            & $Acc_V$ & \textbf{BG}$\downarrow$ & \textbf{VG}$\uparrow$
            & $Acc_V$ & \textbf{BG}$\downarrow$ & \textbf{VG}$\uparrow$
            & $Acc_V$ & \textbf{BG}$\downarrow$ & \textbf{VG}$\uparrow$
            & $Acc_V$ & \textbf{BG}$\downarrow$ & \textbf{VG}$\uparrow$ \\
        \midrule
        Base        & $47.86$ & $34.83$ & $-6.97$ & $45.81$ & $19.35$ & $ 6.46$ & $60.92$ & $19.40$ & $16.52$ & $67.39$ & $12.21$ & $30.18$ & $43.53$ & $ 6.52$ & $12.01$ & $35.18$ & $ 5.26$ & $ 4.68$ \\
        \midrule
        T-SFT       & $50.92$ & $34.87$ & $-3.95$ & $47.37$ & $24.02$ & $ 3.35$ & $67.17$ & $20.28$ & $21.89$ & $72.63$ & $20.28$ & $27.35$ & $47.60$ & $4.53$ & $18.07$ & $35.78$ & $ 6.92$ & $ 3.62$ \\
        T-SFT-DPO   & $51.16$ & $34.59$ & $-3.43$ & $46.41$ & $19.47$ & $ 6.94$ & $67.28$ & $20.67$ & $21.61$ & $67.61$ & $12.52$ & $30.09$ & $47.64$ & $ 4.78$ & $17.86$ & $35.61$ & $ 6.53$ & $ 3.84$ \\
    \bottomrule
    \end{tabular}%
    }
    
    \label{tab:training_results_diag}
\end{table}

\subsection{Threshold Selection and Filtering Results} \label{app:filtering}

Figure~\ref{fig:sweeps} shows the full sweep across all three
signals for both datasets. On MM-AU, $T$ filtering cannot recover positive VG at any threshold, the shortcut structure is too pervasive. The $-V$ sweep improves VG sharply but only at the cost of collapsing the sample count, and increasing BG. The $S$ sweep is the only signal that simultaneously reduces BG and brings VG above zero, with $\tau = 0.1$ falling in a stable region before sample size becomes unreliable. The pattern on VRU-Accident is similar: $T$ suppresses both BG and
VG together, $-V$ overcorrects on VG while raising BG, and $S$ provides the cleanest trade-off.

Table~\ref{tab:dataset_results_filtered} reports the full per-model results before and after filtering. On MM-AU, BG drops from the 25--40 range to 10--19 across all VLMs, VG turns positive for every model, and all three text-only LLMs fall to near or below random chance. On VRU-Accident, BG is roughly halved and VG nearly doubles.

\subsection{Training Evaluation} \label{app:training}

We fine-tune Qwen2.5-VL-7B with supervised fine-tuning on traffic accident data
(T-SFT) and add a DPO stage (T-SFT-DPO), evaluating on traffic accident and general VideoQA benchmarks. Results are in
Tables~\ref{tab:training_results_diag} and~\ref{tab:general_training}.

On MM-AU, training improves video accuracy but BG stays flat and VG remains
negative. On the grounded subset, SFT worsens the grounding profile by raising blind performance alongside video accuracy; DPO largely recovers the base-model grounding on both subsets, suggesting it partially suppresses the shortcuts introduced by SFT. VRU-Accident follows the same pattern on the grounded subset, while the standard subset shows genuine VG improvement with minimal BG change. TrafficQA shows consistent improvement across all three metrics, which is expected given the domain match with the training data. AccidentBench shows negligible changes throughout.

On general VideoQA, VideoMME shows improved BG and VG after training, an
unexpected result suggesting that fine-tuning on traffic data did not narrow
visual utilization to the traffic domain. MLVU shows degraded video accuracy and VG, while LongVideoBench remains largely unaffected. Overall, these results show that accuracy gains do not necessarily correspond to improved visual grounding, and that BG and VG provide a more complete picture of how training affects model behavior.

\begin{table}[t]
    \centering
    \small
    \caption{Performance of Qwen2.5-VL-7B across different model configurations
    in (\%), reported as blind accuracy ($Acc_B$), video accuracy ($Acc_V$),
    Blind Gap (BG), and Visual Gain (VG). The lowest Blind Gap and highest
    Visual Gain per dataset are underlined and bold.}
    \resizebox{\textwidth}{!}{%
    \begin{tabular}{l cccc cccc cccc}
        \toprule
        \multirow{2}{*}{\textbf{Model}}
            & \multicolumn{4}{c}{\textbf{LongVideoBench}}
            & \multicolumn{4}{c}{\textbf{VideoMME}}
            & \multicolumn{4}{c}{\textbf{MLVU}} \\
        \cmidrule(lr){2-5} \cmidrule(lr){6-9} \cmidrule(lr){10-13}
            & $Acc_B$ & $Acc_V$ & $BG{\downarrow}$ & $VG{\uparrow}$
            & $Acc_B$ & $Acc_V$ & $BG{\downarrow}$ & $VG{\uparrow}$
            & $Acc_B$ & $Acc_V$ & $BG{\downarrow}$ & $VG{\uparrow}$ \\
        \midrule
        Base
            & $39.34$ & $53.03$ & $18.22$ & $13.69$
            & $39.07$ & $53.59$ & $14.07$ & $14.52$
            & $42.96$ & $54.52$ & $17.96$ & $11.56$ \\
        \midrule
        T-SFT
            & $38.37$ & $53.25$ & $17.25$ & $14.88$
            & $34.07$ & $53.56$ & $9.07$ & $19.49$
            & $41.36$ & $46.75$ & $16.36$ & $5.39$ \\
        T-SFT-DPO
            & $38.52$ & $54.15$ & $17.04$ & $15.63$
            & $33.70$ & $53.48$ & $8.70$ & $19.78$
            & $41.15$ & $46.50$ & $16.15$ & $5.35$ \\
        \bottomrule
    \end{tabular}}\label{tab:general_training}
\end{table}

\section{Limitations}

\textbf{Dataset scope.} Our audit covers four traffic accident VideoQA benchmarks. While the findings are consistent across datasets and corroborated by general VideoQA results, the degree of modality collapse may differ in other domains or question formats, and the conclusions should not be generalized beyond the evaluated benchmarks without further verification. Additionally, $T(q)$ and $V(q)$ rely on a fixed candidate set and do not extend directly to open-ended generation tasks, where correctness cannot be measured via option probabilities.

\textbf{Threshold selection.} The threshold $\tau = 0.1$ is selected by sweeping BG and VG on the target datasets and using TrafficQA as a grounding reference. This process is dataset-specific and not guaranteed to generalize; a different reference benchmark or a different model panel may yield a different operating
point.

\textbf{Frame sampling and prompting.} All evaluations use 16 uniformly sampled frames and a fixed prompt format. Option ordering, instruction phrasing, and the presence or absence of chain-of-thought prompting can affect output distributions in ways that interact with the shortcut detection methodology.

\textbf{Dependence on model probabilities.} Both T(q) and V(q) rely on output probability distributions, which vary across model architectures and calibration. The Shortcut Score is therefore model-panel dependent: a different set of LLMs or VLMs may assign different scores to the same question. We partially mitigate this by using architecturally diverse panels and evaluating on a held-out model, but the scores are not architecture-agnostic. Cross-family validation with larger-scale models, or human annotations of which questions genuinely require visual input, would provide a stronger model-independent reference.

\textbf{Risk of filtering for model weaknesses.}  Retaining questions where video shifts model behavior toward the correct answer means that the grounding signal is tied to the current panel's visual capabilities. A question may score as visually necessary because the panel responds to its visual content, not necessarily because it requires visual reasoning in principle. Future work with human grounding annotations could provide a model-independent reference.

\end{document}